\documentclass[conference]{IEEEtran}
\IEEEoverridecommandlockouts
\usepackage{cite}
\usepackage{amsmath,amssymb,amsfonts}
\usepackage{listings}
\usepackage{algorithmic}
\usepackage{graphicx}
\usepackage{textcomp}
\usepackage{xcolor}
\usepackage{url}
\usepackage{verbatim}
\usepackage{hyperref}
\usepackage{graphicx}  
\usepackage{adjustbox} 
\def\BibTeX{{\rm B\kern-.05em{\sc i\kern-.025em b}\kern-.08em
    T\kern-.1667em\lower.7ex\hbox{E}\kern-.125emX}}
    
    \definecolor{lightGray}{gray}{0.9}
\begin{document}

\newcommand{\nb}[2]{
    \fbox{\bfseries\sffamily\scriptsize#1}
    {\sf\small\textcolor{red}{\textit{#2}}}
}
\newcommand\ag[1]{\nb{AG}{#1}}
\newcommand\tp[1]{\nb{TP}{#1}}
\newtheorem{puzzle}{Puzzle}

\title{Combining Deep Learning and Explainable AI for Toxicity Prediction of Chemical Compounds}

\author{\IEEEauthorblockN{Eduard Popescu and Adrian Groza 
 \IEEEauthorblockA{\textit{Artificial Intelligence Research Institute - AIRi@UTCN} \\
 \textit{Technical University of Cluj-Napoca},
 Cluj-Napoca, Romania \\
 European University of Technology (EUt+), EU\\
 \tt Adrian.Groza@cs.utcluj.ro}
}
\and
\IEEEauthorblockN{Andreea Cernat
 \IEEEauthorblockA{\textit{Analytical Chemistry Department} \\
 \textit{Iuliu Haţieganu University of Medicine and Pharmacy},\\
 Cluj-Napoca, Romania \\
  \tt acernat.umfcluj.ro}
}}

\maketitle

\begin{abstract}
 The task here is to predict the toxicological activity of chemical compounds based on the  Tox21 dataset, a benchmark in computational toxicology. 
 After a domain-specific overview of chemical toxicity, we discuss current computational strategies, focusing on machine learning and deep learning. 
 Several architectures are compared in terms of performance, robustness, and interpretability. 
 This research introduces a novel image-based pipeline based on DenseNet121, which processes 2D graphical representations of chemical structures. 
 Additionally, we employ Grad-CAM visualizations, an explainable AI technique, to interpret the model’s predictions and highlight molecular regions contributing to toxicity classification. 
 The proposed architecture achieves competitive results compared to traditional models, demonstrating the potential of deep convolutional networks in cheminformatics. Our findings emphasize the value of combining image-based representations with explainable AI methods to improve both predictive accuracy and model transparency in toxicology.
\end{abstract}


\section{Toxicity of chemicals and drugs}

Drug discovery and development is a complex process with continuous monitoring and detailed analysis for every step, that is both time consuming and expensive~\cite{sarvepalli2025ai}. As one of the most transformative technologies, AI can be employed at every stage~\cite{tataru2025ai, cernat2024artificial}, starting from the molecular mechanisms, the available targets, to the development, and efficient analysis of large datasets~\cite{sarvepalli2025ai}. 

De novo drug discovery relied mainly on serendipity and failures are inherent when even small changes can have significant repercussions~\cite{ekins2025neurological}\cite{jusoh2025generative}\cite{fu2025future}. There are billions of possibilities, hence the means to narrow it down to a manageable number to synthesize and test is highly valued. One strategy involves the selection of the targets for different pathologies and then dock millions of molecules into the binding sites for a selection process, while in the absence of a target, a machine-learning model can be generated from the experimental data. In the first case, the identification of the hit molecules is followed by a series of procedures including structures optimization to a achieve a better selectivity, reduced toxicity and improved pharmacokinetic properties \cite{fu2025future}. Preclinical studies that follow after are mainly focused on the mechanism of action and the efficacity in animal models, pharmaceutical formulations and stability tests \cite{jarallah2025ai}. DeepMind AlphaFold2 has an unprecedented high accuracy in blind tests of proteins’ structure prediction to guide experimental NMR data and has greatly improved the number and quality of models of human proteins \cite{ekins2025neurological}\cite{huang2021alphafold2}\cite{casp14website} 

Conventional methods for toxicity in vivo by using animal models can deliver a relative chemical safety profile in human, but this approach is expensive and often difficult to extrapolate the results due to the species variability~\cite{huang2016tox21}. However, the need to minimize in vivo animal toxicity testing, while delivering toxicity data for many chemicals, has fueled the use of advanced technologies to replace these protocols. 
In vitro cytotoxicity assays use different cell lines and can deliver information regarding the capacities of different compounds to destroy the living cells by calculating the cytotoxicity, defined as 50\% inhibition concentration of cell growth. However, by understanding the mechanism of toxicity of several compounds, the process related to cell growth, proliferation and death can be acknowledged~\cite{huang2021toxicity}. 
Even though there is an exponential availability of experimental data regarding chemical or biological compounds, the experimental methods are of high variability and that can generate inconstancies in the public available datasets~\cite{lopez2022sirs}. 

Toxicity prediction models aim to acquire information to display the risk of toxicity by combining chemical toxicity databases with computational techniques. 
Prior 2004, the compounds libraries were a trademark of pharmaceutical companies, having as main purpose the identification of molecules for drug discovery and development, but being corporate investments were not available to the public. Tox21 Program was launched in 2008 as a multi-agency federal partnership (EPA, NTP, and NCGC/NCATS— with FDA joining in 2010)~\cite{richard2021tox21}.  
Tox21 aims to swift evaluation of the possible side effects of the tested compounds and the replacement of the toxicity assessment by mechanism-based toxicity prediction~\cite{jeong2022toxcast}. 


In this research, we predict the toxicity of chemical substances and drugs based on data from the Tox21 dataset
A novel image-based pipeline based on DenseNet121 is introduced to process the 2D graphical representations of chemical structures. 
We also employ Grad-CAM visualizations, an explainable AI technique, to interpret the model’s predictions and highlight molecular regions contributing to toxicity classification. 


\section{Predicting Toxicity Activity}


\subsection{The Tox21 dataset}
The Tox21 Dataset~\cite{tox21}, a publicly available dataset, contains detailed information on the biological effects of a variety of chemical and pharmacological substances. 
It includes data on their effects on different cell types, supporting the researchers to explore the relationship between chemical structure and the toxicological activity of compounds. 
The Tox21 Data Challenge represented the largest community-wide comparison of computational toxicity prediction methods to date. 
It evaluated approximately 12,000 environmental chemicals and pharmaceuticals across 12 high-throughput assays targeting distinct toxicological pathways (Table~\ref{tab:unetloss}, providing a standardized benchmark for model performance~\cite{mayr2016deeptox}. 

\begin{table}
\centering
\caption{Biologic targets}
\label{tab:unetloss}
\begin{tabular}{ll}
Biologic target & Description  \\ \hline
NR-AR    & Androgen receptor \\ 
NR-AR-LBD     & Ligand binding domain of the androgen receptor \\ 
NR-AhR     & Aryl hydrocarbon receptor \\ 
NR-ER     & Estrogen receptor \\ 
NR-ER-LBD     & Ligand binding domain of the estrogen receptor \\ 
NR-PPAR-$\gamma$     & Peroxisome proliferator-activated receptor $\gamma$ \\ 
NR-aromatase     & Aromatase \\ 
SR-ARE     & Antioxidant response element \\ 
SR-ATAD5     & DNA damage response \\ 
SR-HSE     & Heat shock response element \\ 
SR-MMP     & Mitochondrial membrane potential \\ 
SR-p53     & p53 gene activation \\ 
\end{tabular}
\end{table}

\subsection{Predicting toxicity with ML models}
To classify molecules into the 12 biological activity labels, we explored five ML approaches:   
(i) fingerprint-based classical ML models;
(ii) ANN on fingerprints;
(iii) Deep Learning directly on SMILES sequences;
(iv) Graph Neural Networks (GNNs);
(v) image-based feature extraction using DenseNet.
Each approach relies on different types of input representations and model architecture.

\paragraph{Fingerprint-based classical ML models} 
The first approach involved converting SMILES strings into molecular fingerprints, specifically Extended-Connectivity Fingerprints (ECFP4, also known as Morgan fingerprints). 
These fixed-length binary vectors capture the structural characteristics of molecules. Once generated, the fingerprints were used as input features for classical multi-label classification models. Several algorithms were tested, including Random Forests, XGBoost, LightGBM, and Support Vector Machines (SVM). For SVMs, a One-vs-Rest strategy was employed to handle the multi-label nature of the task.

\paragraph{ANNs on Fingerprints}
In a second strategy, the generated fingerprints or molecular descriptors were fed into fully connected neural networks 
These models consisted of simple feedforward dense layers designed to predict the 12 target labels simultaneously. This approach aimed to capture non-linear feature interactions that classical models might miss.

\paragraph{Deep learning directly on SMILES sequences}
Another approach avoided handcrafted features entirely, leveraging the raw SMILES strings as direct input to sequence-based deep learning models. Since SMILES are textual sequences encoding chemical structure, models such as Recurrent Neural Networks (RNNs) — specifically LSTM and GRU architectures — were employed. Additionally, 1D Convolutional Neural Networks (CNNs) were tested to capture local substructure patterns within SMILES. 
More recently, transformer-based models such as ChemBERTa and SMILES transformers provided state-of-the-art performance by learning contextualized embeddings of chemical sequences.

\paragraph{Graph Neural Networks}
Fourth, given the natural graph-like structure of molecules, another method involved converting SMILES into molecular graphs, where atoms are nodes and bonds are edges. Graph Neural Networks (GNNs), specifically Graph Convolutional Networks (GCNs), were employed to perform message passing between nodes, enabling the model to learn representations that incorporate both atomic features and molecular topology. This graph-based learning approach was particularly suited to capturing complex interatomic relationships.

\begin{figure}
    \centering
    \includegraphics[width=0.2\textwidth, height=0.2\textheight, keepaspectratio]{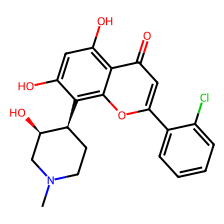}
    \caption{SMILES molecule as image}
    \label{fig:se_out}
\end{figure}

\paragraph{Image-based feature extraction using denseNet}
An alternative technique involved generating 2D images of molecular structures derived from SMILES strings. 
These images (See Figure~\ref{fig:se_out}) were fed into a DenseNet convolutional neural network for feature extraction. 
The high-level features obtained from DenseNet were then used as input for traditional classifiers such as Random Forests, XGBoost, and SVMs to perform the final multi-label classification.
Notably, this approach yielded the best performance across all evaluated metrics, outperforming other methods in terms of both predictive accuracy and robustness. The combination of powerful visual feature extraction with strong tabular classifiers proved to be highly effective for the molecular multi-label classification. 

\subsection{System Architecture}

The proposed architecture capitalizes on the DenseNet feature extraction approach, ensuring a more comprehensive capture of molecular features, and employs XGBoost as the classification model due to its robustness in handling complex, high-dimensional data. The model in Figure~\ref{fig:se_out_1} offers a balance between computational efficiency and prediction accuracy, consisting of:
\begin{figure*}
    \centering
    \includegraphics[width=0.88\textwidth, height=0.8\textheight, keepaspectratio]{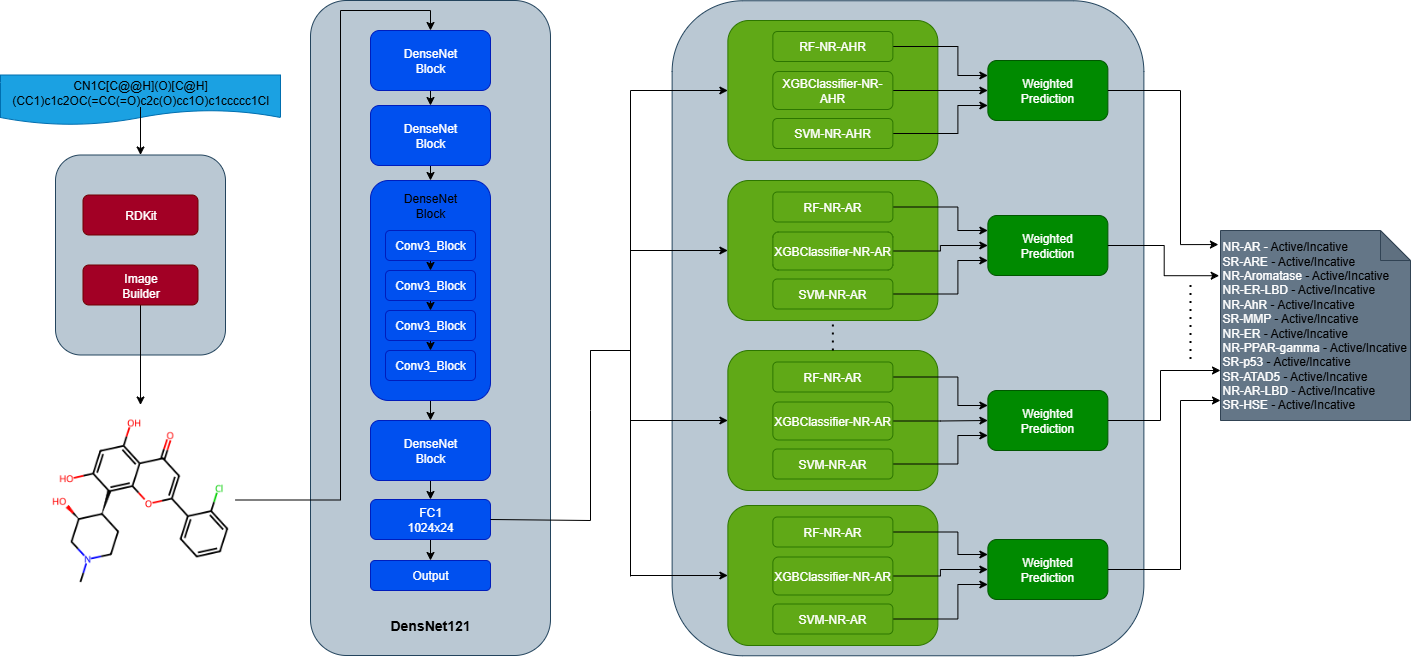}
    \hfill
    \caption{DensNet121 feature extraction and XGBoost-based classification}
    \label{fig:se_out_1}
\end{figure*}

First, the \textit{initial convolution layer} applies a large filter (7$\times$7) to detect basic shapes, such as edges and textures. 
It includes a max pooling operation to reduce resolution.
\begin{equation}
g(x, y) = \sum_{i=-a}^{a} \sum_{j=-b}^{b} f(x+i, y+j) \times k(i, j)
\end{equation}
where: \( f(x, y) \) is the input function, i.e. the image or the feature map,
    \( k(i, j) \) is the kernel (filter) applied,
     \( a = \left\lfloor \frac{m}{2} \right\rfloor, \quad b = \left\lfloor \frac{n}{2} \right\rfloor \), for a kernel of size \( m \times n \),
     \( g(x, y) \) is the resulting pixel at position \((x, y)\) after applying convolution.

Second, \textit{batch normalization} is applied in several key locations in the DenseNet121 architecture. 
Specifically, it appears before each convolution in the Dense and transition blocks. 
In the transition layer, batch normalization is applied to the accumulated output from the previous block, just before reducing the dimensionality via convolution.

\begin{equation}
\hat{x}^{(k)} = \frac{x^{(k)} - \mu^{(k)}}{\sqrt{\sigma^{2(k)} + \epsilon}}, \quad
y^{(k)} = \gamma^{(k)} \hat{x}^{(k)} + \beta^{(k)}
\end{equation}
where: \( x^{(k)} \) is the initial activation for channel \(k\),
    \( \mu^{(k)} \), \( \sigma^{2(k)} \) are the batch mean and variance,
     \( \epsilon \) is a very small constant for numerical stability,
     \( \gamma^{(k)}, \beta^{(k)} \) are learned parameters (scale and shift).

Third, \textit{dense block} is composed of multiple dense layers that work together. 
Each layer receives information from all previous ones, which facilitates learning and avoids information loss. 
DenseNet121 includes 4 such blocks, with 6, 12, 24, and 16 layers respectively.

Fourth, the \textit{dense layer} within a dense block consists of two steps: 
(i) it applies a 1$\times$1 filter to reduce dimensionality and computations; 
(ii) then a 3$\times$3 filter extracts new features. 
It contributes unique information, while relying on everything learned before.

Fifth, the \textit{transition Layer}, found between dense blocks, reduces the data dimensionality and the number of channels, so the model does not grow too large. 
It includes a $1\times 1$ filter and an average pooling operation. 
The transition layer maintains a balance between performance and efficiency, 

Sixth, the \textit{pooling layer} transforms each feature map into a single value by computing the average. 
The goal is to reduce the data to a simple form that the final layer can easily use.
At the end, after the image has passed through all blocks, a layer performs averaging on each feature map to obtain simple values. 
Then, these values are sent to a final layer that decides the image's class. 
Since the network was used as a feature extractor, the classification layer which provides the decision based on extracted features, was omitted.
\begin{equation}
Y_c = \frac{1}{H \cdot W} \sum_{i=1}^{H} \sum_{j=1}^{W} X_c(i,j)
\end{equation}
with $y = W x + b$. 
Here, 
 \( X_c(i, j) \) is the value at position \((i, j)\) in channel \(c\),
     \( H \times W \) is the spatial dimension of the map,
     \( Y_c \) is the computed average for channel \(c\),
     \( x \in \mathbb{R}^n \) is the input vector,
     \( W \in \mathbb{R}^{m \times n} \) is the weight matrix,
     \( b \in \mathbb{R}^m \) is the bias vector,
     \( y \in \mathbb{R}^m \) is the resulting output.

Seventh, the \textit{ML block} receives 1,056 features from the DenseNet for each chemical compound image. 
Three classifiers are used: (i) Support Vector Machine with a linear kernel for its interpretability, (ii) Random Forest for its robustness against overfitting, and 
(iii) XGBoost for its high accuracy and speed through sequential boosting and regularization. 
These models operate independently, and a majority voting strategy is used to combine their predictions. 
Although a weighted voting approach was explored, it didn’t significantly improve performance and was thus excluded. The system offers a reliable way to classify chemical compounds from 2D images while preserving critical structural information.

\section{Running experiments}

\begin{table*}
\centering
\caption{Accuracy per label for each model tested}
\label{tab:label_accuracy}
\begin{tabular}{lcccccccccccc}
\textbf{Model} & \textbf{NR-AR} & \textbf{SR-ARE} & \textbf{NR-Aroma} & \textbf{NR-ER} & \textbf{NR-AhR} & \textbf{SR-MMP} & \textbf{NR-ER} & \textbf{NR-PPAR} & \textbf{SR-p53} & \textbf{SR-ATAD5} & \textbf{NR-LBD} \\
\hline
RF + FP & 0.75 & 0.70 & 0.68 & 0.72 & 0.69 & 0.71 & 0.67 & 0.73 & 0.70 & 0.69 & 0.71 \\
XGBoost + FP & 0.76 & 0.72 & 0.69 & 0.74 & 0.71 & 0.73 & 0.69 & 0.75 & 0.71 & 0.70 & 0.72 \\
MLP + FP & 0.74 & 0.69 & 0.66 & 0.71 & 0.67 & 0.70 & 0.66 & 0.71 & 0.68 & 0.67 & 0.69 \\
RNN on SMILES & 0.72 & 0.68 & 0.65 & 0.79 & 0.76 & 0.78 & 0.74 & 0.70 & 0.77 & 0.75 & 0.78 \\
GNN (GCN) & 0.78 & 0.75 & 0.73 & 0.77 & 0.74 & 0.76 & 0.72 & 0.78 & 0.75 & 0.73 & 0.76 \\
DenseNet + RF & 0.92 & 0.91 & 0.94 & 0.89 & 0.93 & 0.90 & 0.89 & 0.83 & 0.93 & 0.94 & 0.95 \\
DenseNet + XGBoost & 0.93 & 0.91 & 0.94 & 0.88 & 0.93 & 0.91 & 0.88 & 0.84 & 0.93 & 0.94 & 0.93 \\
DenseNet + SVM & 0.94 & 0.91 & 0.94 & 0.90 & 0.93 & 0.90 & 0.90 & 0.84 & 0.94 & 0.94 & 0.96 \\
\hline
\end{tabular}
\end{table*}

\begin{table}
\centering
\label{tab:label_accuracy_1}
\begin{adjustbox}{valign=c}  
\begin{tabular}{l@{\hskip 5pt}cccccccccccc}
\textbf{Model} &  \textbf{SR-HSE} \\
\hline
RF + FP & 0.68 \\
XGBoost + FP & 0.69 \\
MLP + FP & 0.66 \\
RNN on SMILES & 0.75 \\
GNN (GCN) & 0.74 \\
DenseNet + RF & 0.91 \\
DenseNet + XGBoost & 0.89 \\
DenseNet + SVM & 0.92 \\
\hline
\end{tabular}
\end{adjustbox}
\end{table}

\begin{table}
\caption{Results on DenseNet121 with the ensemble of SVM, XGBoost, Random Forest}
\centering
\begin{tabular}{ll}
\textbf{Label} & \textbf{Accuracy} \\ \hline
NR-AHR               & 85.90 \\ 
NR-AR              & 79.67\\ 
AR-LBD              & 90.12 \\ 
NR-Aromatase              & 91.67\\ 
NR-ER               & 84.41 \\ 
NR-ER-LBD              & 79.67\\ 
NR-PPAR-GAMMA              & 90.12 \\ 
SR-ARE              & 91.32\\ 
SR-ATAD5               & 88.41 \\ 
SR-HSE              & 89.67\\ 
SR-MMP              & 90.12 \\ 
SR-P53              & 86.67\\ 
\end{tabular}
\label{tab:tabPF}
\end{table}

We present the results for the toxicity activity prediction task, which aims to classify chemical compounds as active or inactive against specific biological targets. 
Several deep learning architectures were evaluated, including ResNet50, EfficientNetB0, and InceptionV3, each offering different trade-offs between performance and computational cost. Among these, the proposed architecture based on DenseNet121 demonstrated superior generalization capabilities, particularly when processing 2D representations of chemical structures (see Tables~\ref{tab:label_accuracy},~\ref{tab:tabPF}).

\begin{figure*}
    \centering
    \includegraphics[width=0.24\textwidth, height=0.16\textheight]{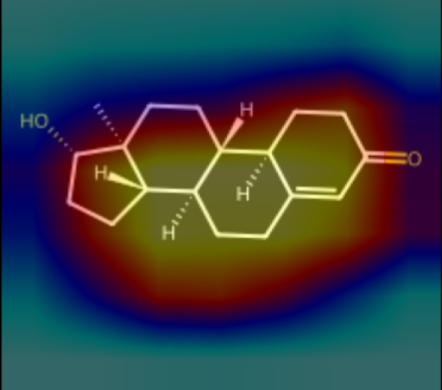}\hfill
    \includegraphics[width=0.24\textwidth, height=0.16\textheight]{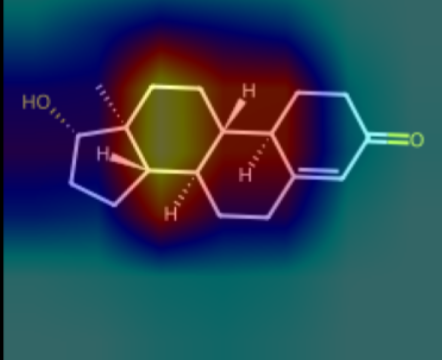}\hfill
    \includegraphics[width=0.24\textwidth, height=0.16\textheight]{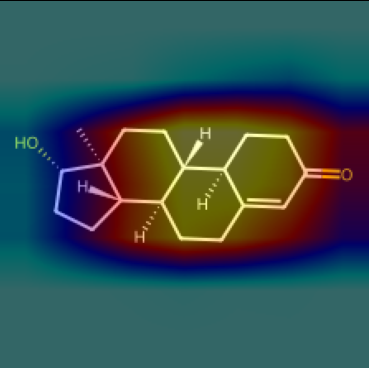}
    
    \caption{Grad-CAM image highlighting the region of interest in the decision-making for the NR-AR label (left), NR-ER (center) and SR-P53 (right) }
    \label{fig:imgGradNR-AR}
\end{figure*}

Unlike traditional descriptor-based models, our approach relies entirely on visual features extracted from the structural images of compounds. To provide interpretability and insight into the model's decision-making process, we applied Grad-CAM visualizations (Figure~\ref{fig:imgGradNR-AR}). 
These heatmaps highlight the regions of the input images that contributed most significantly to each classification decision, offering a more transparent and biologically plausible understanding of the predictions.

For each of these labels, the report indicates whether the molecule was classified as active or inactive, along with the corresponding probability of the outcome.
To assess the robustness and reliability of the generated result, a global confidence score of the prediction is calculated, consisting of two components.

First, the \textit{confidence of the deep learning model (DenseNet121)} – derived from the variability in feature extraction, obtained through multiple stochastic runs:
    \begin{equation}
    Trust_{\text{DenseNet}} = 1 - \frac{1}{D} \sum_{i=1}^{D} \sigma_i
    \end{equation}
    where $\sigma_i$ represents the standard deviation of each dimension $i$ from the feature vector of dimension $D$.

Second, the \textit{confidence score of the ML block} is calculated by aggregating the probability scores returned by each component algorithm, weighted by their importance:
\begin{equation}
Trust_{\text{ML}} = \sum_{i=1}^{n} \alpha_i \cdot p_i
    \end{equation}
where $p_i$ is the probability provided by the ML model $i$ (i.e. SVM, Random Forest, or XGBoost), and $\alpha_i$ is its weight in the decision-making process.

The combination of these two components enables the formulation of a final confidence score for each prediction, providing the user with an objective measure of the stability and certainty of the automated decision. 
This approach enhances transparency and supports the interpretability of results~\cite{wu2021trade} in the context of computational toxicology applications.

\section{Discussion and related work}
Chemical toxicity is a key factor in assessing the safety and effectiveness of chemical substances and pharmaceuticals, having a significant impact on human health, the environment, and the pharmaceutical industry. In the fast-paced context of chemical and pharmaceutical discovery and use, the rapid and accurate identification of their potential toxic effects becomes essential. Currently, traditional toxicity testing methods are costly, time-consuming, and often involve animal use, which raises ethical and economic concerns. In this context, the use of advanced machine learning technologies for toxicity prediction is becoming a promising and efficient alternative.

Chemical toxicity matters when we judge how safe and effective chemicals and drugs are. 
It affects people’s health, the environment, and the drug industry. Today, new chemicals and medicines are found and used quickly, so we must spot possible toxic effects fast and accurately. 
Traditional tests are slow, expensive, and often use animals, which raises ethical and cost concerns.
In this setting, artificial intelligence offers an alternative for predicting toxicity—faster, cheaper~\cite{mustafa2025tabnet}, and with less animal testing.

In computational toxicology, recent research has focused on the application of ML to predict the toxicity of chemical substances and drugs. The use of public datasets such as Tox21 has become a relevant starting point for developing predictive models, as these datasets offer detailed information on the biological effects of a large number of chemical compounds.
Chen et al.~\cite{chen2016tox21} has applied deep learning on the Tox21 dataset. The approach was followed by researchers who combined chemical structure analysis with toxicological modeling to predict effects on different cell types.

An important branch of toxicology research is the evaluation of chemical purity. 
The purity of a chemical compound can directly influence its toxicological activity, hence this aspect is often addressed in virtual screening research. In this line, Köhler et al. \cite{koehler2018toxicity} have used Tox21 to evaluate the purity and toxicological activity of chemical compounds, implementing purity ranking algorithms based on chemical structure and observed biological effects.

In the field of drug discovery, Hawkins et al.~\cite{hawkins2020drug} have emphasized the evaluation of pharmaceutical drug purity and their interactions with various biological targets. They showed that impurities and contaminants can significantly alter a drug’s effectiveness and toxicity profile, making it essential to develop models capable of ranking drug purity based on toxicity and biological activity.

Zhang et al.~\cite{zhang2021multi} used multi-task learning techniques to simultaneously predict multiple aspects of toxicity and chemical purity. The multi-task model has enabled both toxicity prediction and also the classification of chemical compounds based on their relative purity. 

Liu et al.~\cite{liu2019transfer} have applied transfer learning on Tox21 to improve the accuracy of toxicological predictions for new chemical substances, which is an essential step when aiming to rank the purity of a broad set of unknown chemicals.

Various ML have been applied for predicting toxicity and purity, including XGBoost, LightGBM, Random Forests or CNNs. 
Liu et al.~\cite{liu2018xgboost}, XGBoost models were used to predict the toxic risks of chemical substances in Tox21, with promising results in terms of prediction accuracy.
Zhou et al.~\cite{zhou2020cnn} have used CNNs to analyze the spectroscopic data of chemical compounds, thus demonstrating the applicability of deep learning methods and chemical data analysis in toxicity prediction.

For drug discovery, Tox21 was used for accelerating the screening and selection of compounds. Chakravarti et al.~\cite{chakravarti2017toxicity} have combined chemical and biological data analysis to predict drug toxicity. 
The learned models were able to: 
(i) efficiently distinguish between safe and toxic-risk drugs, and 
(ii) identify chemical substances that could interact with various biological targets. 
The solution  provides a risk assessment and classification of substance purity based on their toxicological potential.


\section*{Acknowledgment}
This work is supported by the project "Romanian Hub for Artificial Intelligence-HRIA", Smart Growth, Digitization and Financial Instruments Program, MySMIS no. 334906

\section{Conclusion}


This study explored the application of machine learning and deep learning  for toxicity prediction using the Tox21 dataset. After providing a domain-specific introduction to chemical toxicity and computational approaches, we analyzed the performance of several existing architectures and compared them with a proposed DenseNet121-based model. The proposed approach leverages image representations of chemical structures and Grad-CAM visualization for interpretability.  
These results highlight the potential of image-based deep learning models in predictive toxicology and support further development in this direction.

\bibliographystyle{IEEEtran}
\bibliography{bib} 

\end{document}